# Rotation Invariant Face Detection Using Wavelet, PCA and Radial Basis Function Networks

S. M. Kamruzzaman[1], Firoz Ahmed Siddiqi[2], Md. Saiful Islam[3],
Md. Emdadul Haque[1] and Mohammad Shamsul Alam[2]

[1]Department of Information and Communication Engineering
[3]Department of Computer Science and Engineering
University of Rajshahi, Rajshahi-6205, Bangladesh.
[2]Department of Computer Science and Engineering
International Islamic University Chittagong, Dhaka Campus, Bangladesh.
E-mail: smzaman@gmail.com, {firoz_moni, emdad_74, alam_cse}@yahoo.com,
msis_ru@yahoo.co.in

**Abstract.** This paper introduces a novel method for human face detection with its orientation by using wavelet, principle component analysis (PCA) and redial basis networks. The input image is analyzed by two-dimensional wavelet and a two-dimensional stationary wavelet. The common goals concern are the image clearance and simplification, which are parts of de-noising or compression. We applied an effective procedure to reduce the dimension of the input vectors using PCA. Radial Basis Function (RBF) neural network is then used as a function approximation network to detect where either the input image is contained a face or not and if there is a face exists then tell about its orientation. We will show how RBF can perform well then back-propagation algorithm and give some solution for better regularization of the RBF (GRNN) network. Compared with the traditional RBF networks, the proposed network demonstrates better capability of approximation to underlying functions, faster learning speed, better size of network, and high robustness to outliers.

## 1. Introduction

Face recognition has drawn considerable interest and attention from many researchers in the pattern recognition field for the last few decades [2]. The recognition of faces is very important because of its potential commercial applications, such as in the area of video surveillance, robotic control system, access control systems, retrieval of an identity from a data base for criminal investigations and user authentication[3]. One of typical procedures can be described for video surveillance applications. A system that automatically recognizes a face in a video stream first detects the location of face and normalizes it with respect to the pose, lighting and scale. Then, the system tries to extract some pertinent features and to associate the face to one or more faces stored in its database, and gives the set of faces that are considered as nearest to the detected face [4]. Usually, each of these stages for detection, normalization, feature extraction and recognition is so complex that it must be studied separately [1].

In this paper we have developed a real time face detector with rotation indicator, which is robust in different conditions for embedded systems. Face detection is a difficult task in image analysis, which has each day more and more applications. The existing methods for face detection can be divided into image based methods and feature based methods. Our developed system is an intermediate system used a radial basis function network to train the classifier, which is capable of processing images rapidly while having high detection rates.

## 2. Wavelet and De-Noising Procedure Principle

Two-dimensional wavelet analysis and a two-dimensional stationary wavelet analysis are used to de-noise an image. De-noising is one of the most important applications of wavelets [5]. The general de-noising procedure involves three steps. The basic version of the procedure follows the steps described below.

- Decompose: Choose a wavelet from different kind; choose a level N. Compute the wavelet decomposition of the signals at level N.
- Threshold Detail Coefficients: For each level from 1 to N, select a threshold and apply soft threshold to the detail coefficients.
- Reconstruct: Compute wavelet reconstruction using the original approximation coefficients of level N and the modified detail coefficients of levels from 1 to N.

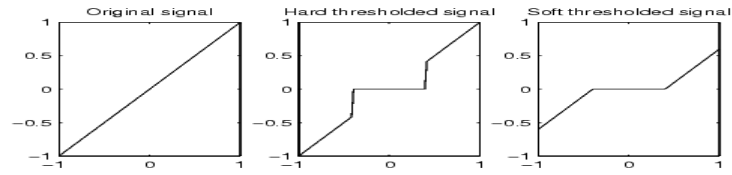

**Fig.1.** Soft and Hard threshold.

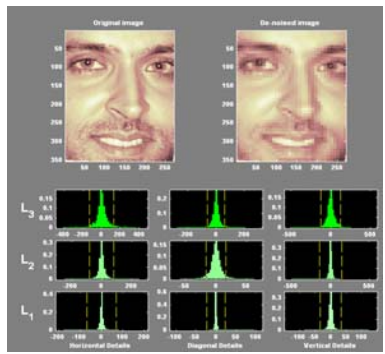 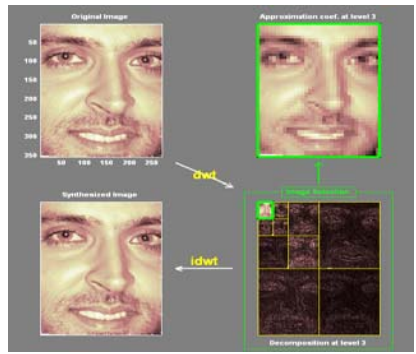

**Fig. 2.** Level 3 analysis of decomposition.   **Fig. 3.** Choosing soft threshold and de-noised.

Hard threshold is the simplest method but Soft threshold has nice mathematical properties [6]. Soft threshold is an extension of hard threshold, first setting to zero the elements whose absolute values are lower than the threshold, and then shrinking the nonzero coefficients towards 0.

## 3. Principal Component Analysis

Principal Components Analysis (PCA), also known as Karhunen-Loeve methods, is a technique now commonly used for dimensionality reduction in computer vision, particularly in face recognition [7]. A method called Eigenface based on PCA was used in face recognition in which chooses a dimensionality reducing linear projection that maximizes the scatter of all projected samples.

## 4. Training GRNN with Sigmoidal Radial Basis Function

We manually collect 30 different types of images with different illumination. First every image is resizing with [15*15] size. The size is choice by a small training contributing only 5 images. The training result is plot bellow. The vertical line shows the resizing square matrix. At size [15*15] the error is little and after that error is almost constant.

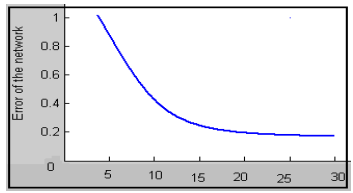
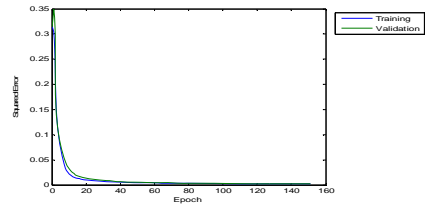

**Fig. 4.** Training error.                **Fig. 5.** Rprop training.

## 5. Algorithm for Searching Face

The following algorithm is used to detect the face in any image:

> *Let f(x,y) is an gray image*
> *where x=1,…,N;*
> *and y=1,…,M;*
> ***1.*** *for i=1 to N-15*
> ***2.*** *   for j=1 to m-15*
> ***3.*** *      crop image d(k,l)*
> *                   where x=1,…,15;*

         *and y=1,…,15;*
  **4.**    *Apply 2D wavelet for de-noise*
  **5.**    *Apply PCA*
  **6.**    *Simulate by* GRNN
  **7.**    *If Simulate value with in 90 to-90*
        *Marking with degree;*
     *else*
       *continue;*

## 6. Reducing Training Time by GRNN

The training is running almost 150 epochs still it does not reach to target. The training is carried out on our input patterns and takes almost 3 hours to achieve error of 0.01. But in GRNN there need only a few second to train the network.

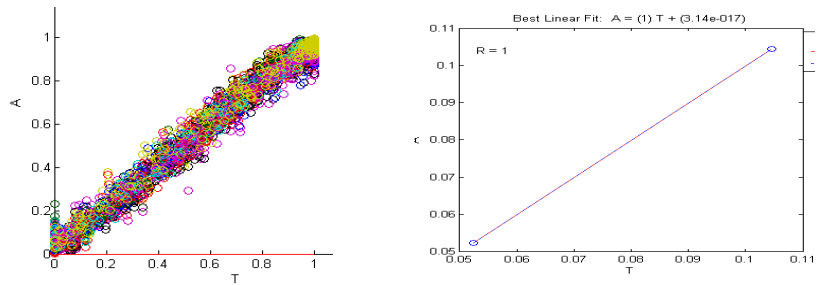

(a) m= 0.9707 b= 0.0073 r= 0.9995 (b) m = 1.0000 b = 3.1402e-017 r =1.0000

**Fig. 6.** Miss classifying rate reduce (a) For Rprop (b) For GRNN

The first two, m and b, correspond to the slope and the y-intercept of the best linear regression relating targets to network outputs. If this number is equal to 1, then there is perfect correlation between targets and outputs. In our example, the number is m and r =1 and b = 0 for Fig. 6(b), which indicates a good fit. If we closely see the Fig. 6(a) then it is clear that 20% of data sets are miss classified. On the other hand for Fig. 6(b) misclassifying is almost is zero.

## 7. Comparative Study

Table 1 shows a breakdown of the detection rates of different systems on faces that are rotated less and more than$10^0$ from uptight. In the table we compare our proposed method with other method and showed that it works very good compeering with others. Its computation speed is also as fast as Viola-Jones. Bellow we give some real world examples. It has been seen from Fig. 7 that the proposed face detector is able to detect faces on a number of test images containing complex background and also tells the orientation correctly.

**Table 1.** Breakdown of the detection rates of different systems.

| Detector | All Faces | Upright Faces ($<=10^0$) | Rotated Faces ($>10^0$) |
|---|---|---|---|
| Viola-Jones | 92.1% | 93.0% | 94.1% |
| Rowley-Baluja-Kanade | - | 90.1% | 89.9% |
| Proposed method | 93.01% | 94.0% | 94.0% |

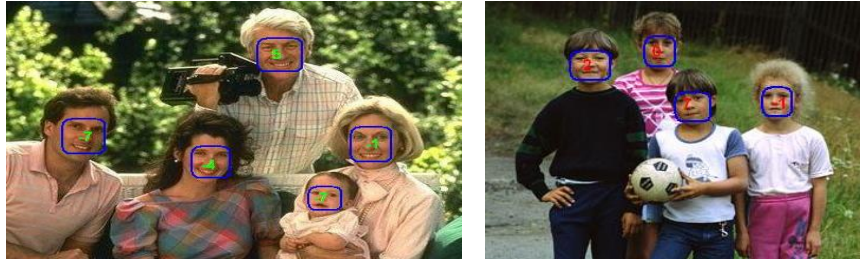

**Fig. 7.** Output of the proposed face detector.

## 8. Conclusions and Further Work

We have presented an approach for face detection, which minimizes computation time while achieving high detection accuracy. The approach performs well compared with existing systems for its fast detecting and also tells its orientation. Though its work well for frontal and semi-profile images, it has less recognition capability on profile face. The contribution of this paper is a simple and efficient classifier, which is computationally efficient. This classifier is clearly an effective one for face detection and we are confident that it will also be effective in other domains such as automobile or pedestrian detection.